\documentclass[final]{cvpr}

\usepackage{times}
\usepackage{epsfig}
\usepackage{graphicx}
\usepackage{amsmath}
\usepackage{amssymb}
\usepackage{multirow}
\usepackage[ruled]{algorithm2e}
\usepackage{float}
\usepackage{subfigure}
\usepackage[pagebackref=true,breaklinks=true,colorlinks,bookmarks=false]{hyperref}

\def\confYear{CVPR 2021}
\usepackage{pifont}

\begin{document}

\title{Shallow Feature Matters for Weakly Supervised Object Localization}

\author{Jun Wei\textsuperscript{1,2,\rm $\dagger$} \quad
        Qin Wang \textsuperscript{1,2,\rm $\dagger$}\quad
        Zhen Li$^{1,2,}$\thanks{{Corresponding author. $^\dagger$ Equal contribution.}}\quad
        Sheng Wang$^5$ \quad 
        S.Kevin Zhou$^{4,3,1}$ \quad 
        Shuguang Cui$^{1,2}$ \\
        $^1$ School of Science and Engineering, The Chinese University of Hong Kong (Shenzhen) \\
	    $^2$ Shenzhen Research Institute of Big Data \quad
	    $^3$ University of Science and Technology of China \\
	    $^4$ Institute of Computing Technology, Chinese Academy of Sciences \quad
	    $^5$ CryoEM Center, SUSTech \\
{\tt\small	\{junwei@link., qinwang1@link., lizhen@\}cuhk.edu.cn, s.kevin.zhou@gmail.com}}

\maketitle

\begin{abstract}
Weakly supervised object localization (WSOL) aims to localize objects by only utilizing image-level labels. Class activation maps (CAMs) are the commonly used features to achieve WSOL. However, previous CAM-based methods did not take full advantage of the shallow features, despite their importance for WSOL. Because shallow features are easily buried in background noise through conventional fusion. In this paper, we propose a simple but effective {\textbf{S}}hallow feature-aware {\textbf{P}}seudo supervised {\textbf{O}}bject {\textbf{L}}ocalization ({\textbf{SPOL}}) model for accurate WSOL, which makes the utmost of low-level features embedded in shallow layers. In practice, our SPOL model first generates the CAMs through a novel element-wise {\textbf{multiplication}} of shallow and deep feature maps, which filters the background noise and generates sharper boundaries robustly. Besides, we further propose a general class-agnostic segmentation model to achieve the accurate object mask, by only using the initial CAMs as the pseudo label without any extra annotation. Eventually, a bounding box extractor is applied to the object mask to locate the target. Experiments verify that our SPOL outperforms the state-of-the-art on both CUB-200 and ImageNet-1K benchmarks, achieving 93.44\% and 67.15\% ({\it i.e.,} {\textbf{3.93\%}} and {\textbf{2.13\%}} improvement) Top-5 localization accuracy, respectively.
\end{abstract}

\section{Introduction}
Weakly supervised object localization (WSOL) aims to locate objects by using only image-level labels. 
Since no expensive bounding box annotations are required, WSOL has attracted lots of attentions in various applications, \eg lesion localization for medical image diagnosis, image-label guided retrieval, etc~\cite{camcvpr2016,acolcvpr2018,adlcvpr2019,aecvpr2017,bae2020rethinking,psolcvpr2020,cutmixiccv2019,pairwiseeccv2020,hideandseekiccv2017,spgeccv2018,daneticcv2019,eilcvpr2020,Zhang0YWZH20}.
\begin{figure}[t]
    \centering
    \includegraphics[width=\columnwidth]{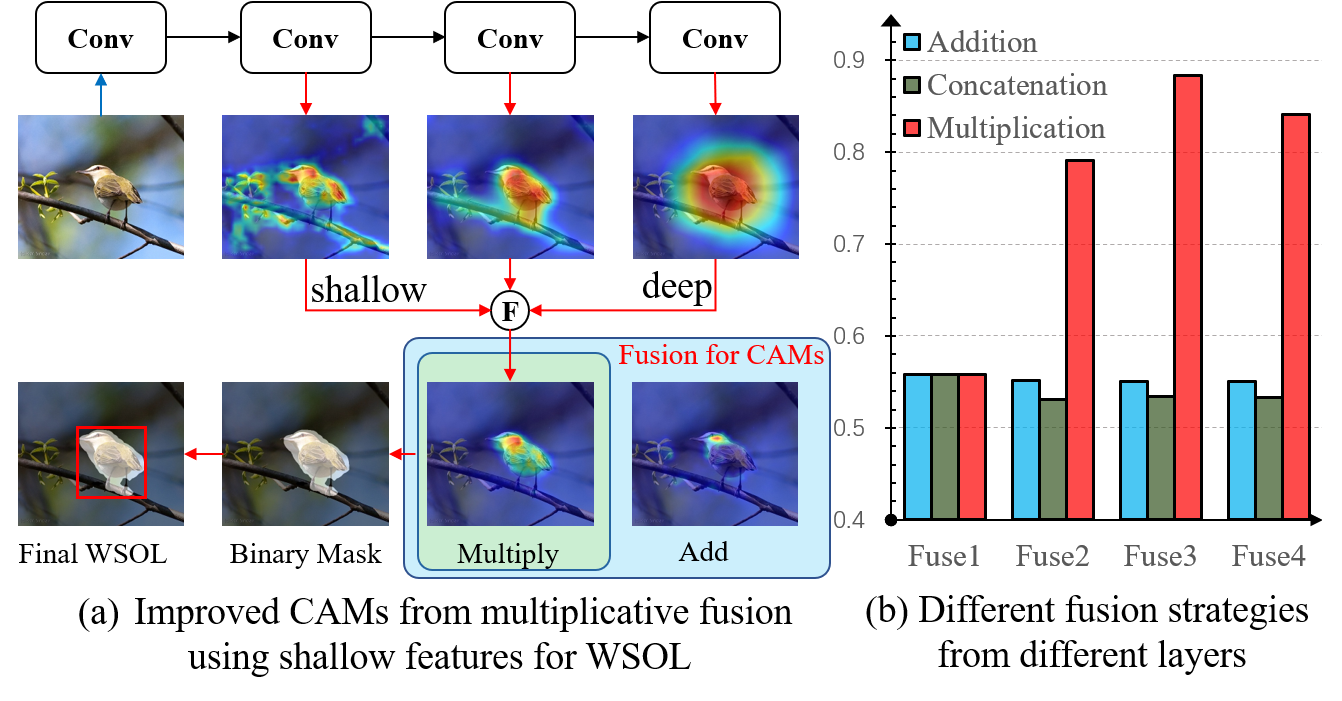}
    \caption{(a) CAM-based pipeline for Weakly Supervised Object Localization (WSOL). $\mathcal{F}$ represents different fusion strategies. Here multiplication and addition based fusion methods for CAMs are compared. (b) \textit{GT-known Loc} comparison using fused CAMs through different fusion strategies from different layers. In practice, Fuse$x$ means the last $X$ stage features of ResNet50 are aggregated, \eg Fuse1 represents the original CAMs.}
    \label{fig:teaser:framework}
\end{figure}

\begin{figure}[t]
    \centering
    \includegraphics[scale=0.53]{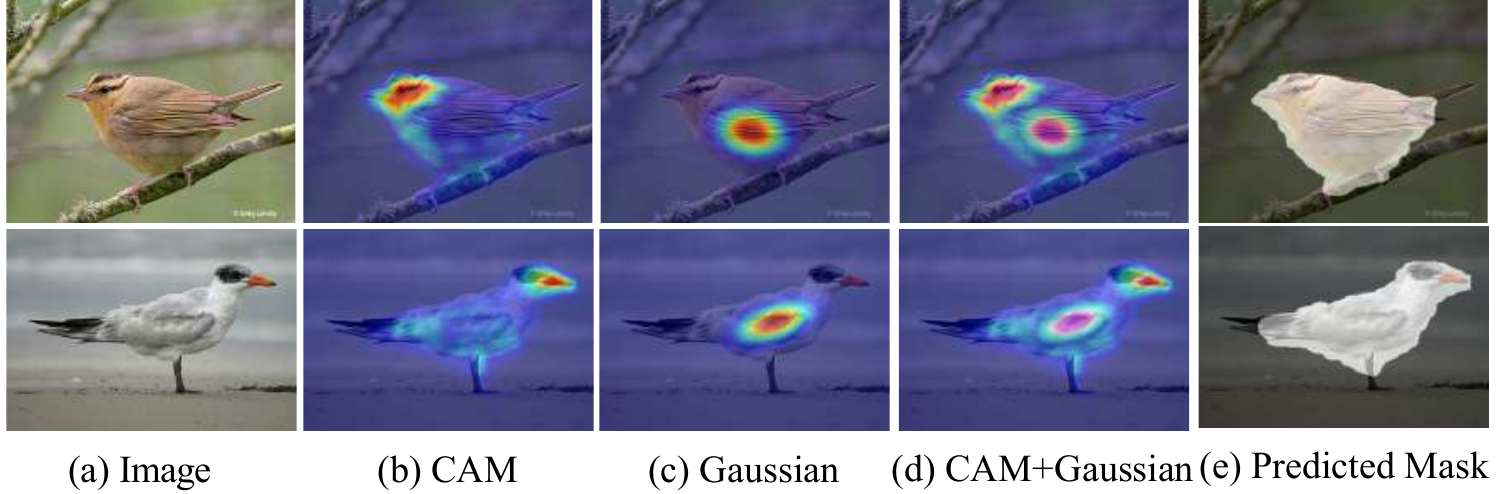}
    \caption{Variants of activated regions. (a) Input image. (b) Original CAM~\cite{camcvpr2016}. (c) Gaussian prior pseudo labels using original CAM as the weighting coefficients. (d) Gaussian prior pseudo labels combined with original CAM. (e) Object mask predicted by the proposed class-agnostic segmentation model.}
    \label{fig:teaser:guassian}
\end{figure}

Existing WSOL methods are mainly based on the class activation maps (CAMs)~\cite{camcvpr2016}.
However, CAM-based models are initially trained for classification, where the network objective is inconsistent with localization. 
%
%
Specifically, classification prefers features with semantic meanings, usually derived from deep layers of convolutional neural networks (CNNs). 
%
%
In contrast, shallow features, derived from the shallow layers of CNNs, have less semantics but are rich in details, which have clearer edges and less distortion.
%
Unfortunately, direct fusion of shallow and deep features is invalid for WSOL due to the following two defects:
1) The meaningful information embedded in the shallow features cannot be well captured through weak supervision, due to the large interference of background noise. 
%
%
To better elaborate this statement, we illustrate the CAM-based WSOL pipeline in Fig.~\ref{fig:teaser:framework} (a).
Within this pipeline, features of different layers from ResNet50~\cite{resnetcvpr2016} are firstly aggregated to form the fused CAMs.
As the fused CAMs obtained through addition operation shown in Fig.~\ref{fig:teaser:framework} (a), the shallow features are buried and not fully utilized, leading to inferior CAMs.
%
To make it general, as shown in Fig.~\ref{fig:teaser:framework} (b), we further test other conventional CAM fusion strategies (\eg concatenation) for different layers, which are evaluated by \textit{GT-known Loc} on CUB-200~\cite{cubtech2011} dataset.
Regarding the quantitative and qualitative results, neither the obtained CAMs nor the overall localization accuracy has been improved when more shallow features are involved through conventional fusion strategies, \ie addition or concatenation.
2) Another issue is that only the most discriminative regions are activated in the original CAMs~\cite{acolcvpr2018,aecvpr2017,hideandseekiccv2017,cutmixiccv2019,eilcvpr2020,adlcvpr2019}. 
As shown in Fig.~\ref{fig:teaser:framework} (a) and Fig.~\ref{fig:teaser:guassian} (b), most areas have low response except for the head region, even though the low response areas occupy most of the image and reflect the object shape.

To address above concerns, we propose a simple but effective {\textbf{S}}hallow feature-aware {\textbf{P}}seudo supervised {\textbf{O}}bject {\textbf{L}}ocalization (named SPOL) model for accurate WSOL, which makes the utmost of the low-level features embedded in shallow layers. 
Our SPOL model mainly consists of two stages, \ie CAM generation and class-agnostic segmentation. 
For the CAM generation, the multiplicative feature fusion network (MFF-Net) is designed to aggregate both shallow and deep features. 
Different from previous fusion methods, features in MFF-Net are treated in a synergistic way. 
Namely, deep features with clear background help suppress the noise of the shallow ones while shallow features with rich local structures make object boundaries sharper, just as the multiplicative fusion CAMs shown in Fig.~\ref{fig:teaser:framework} (a) and performance gains shown in Fig.~\ref{fig:teaser:framework} (b).
For the class-agnostic segmentation stage, initial CAMs will be refined with the Gaussian prior pseudo label (GPPL) module, which is then regarded as the pseudo label for class-agnostic segmentation module training.
Specifically, taking full advantage of the initial entire CAM as the weighting coefficients, the mean and variance for all coordinates are calculated to obtain the object gravity.
Then, a Gaussian distribution can be generated with achieved mean and variance, called Gaussian prior pseudo label (GPPL).
As shown in Fig.~\ref{fig:teaser:guassian} (c), GPPL approximates the gravity center of the bird, and enhances the responses of areas inside the body. 
Combining GPPL and original CAM, a better CAM can be obtained, just as Fig.~\ref{fig:teaser:guassian} (d) shows. 
To further refine these regions, we design a class-agnostic segmentation model by using combined GPPL and CAMs as pseudo labels through another MFF-Net.
Note that during the training phase, areas with large and small CAMs responses will be binarized into foreground and background, respectively using two pre-defined thresholds, and other parts will be ignored to avoid label conflict during training. 
After training, the obtained object mask will become more complete compared with initial CAMs, as shown in Fig.~\ref{fig:teaser:guassian} (e).
Finally, a bounding box extractor is applied to the object mask to obtain the final object localization. In summary, our contributions are three-fold:
\begin{itemize}
    \item We propose a SPOL model to fully utilize the vital shallow features for WSOL, owing to the proposed multiplicative feature fusion strategy that makes the utmost of shallow features.
    
    \item We further propose the Gaussian prior pseudo label (GPPL) and class-agnostic segmentation model to achieve a better object mask for WSOL.

    \item SPOL outperforms previous methods by a large margin on both CUB-200 and ImageNet-1K benchmarks.
\end{itemize}

\begin{figure*}[t]
    \centering
    \includegraphics[width=0.8\textwidth]{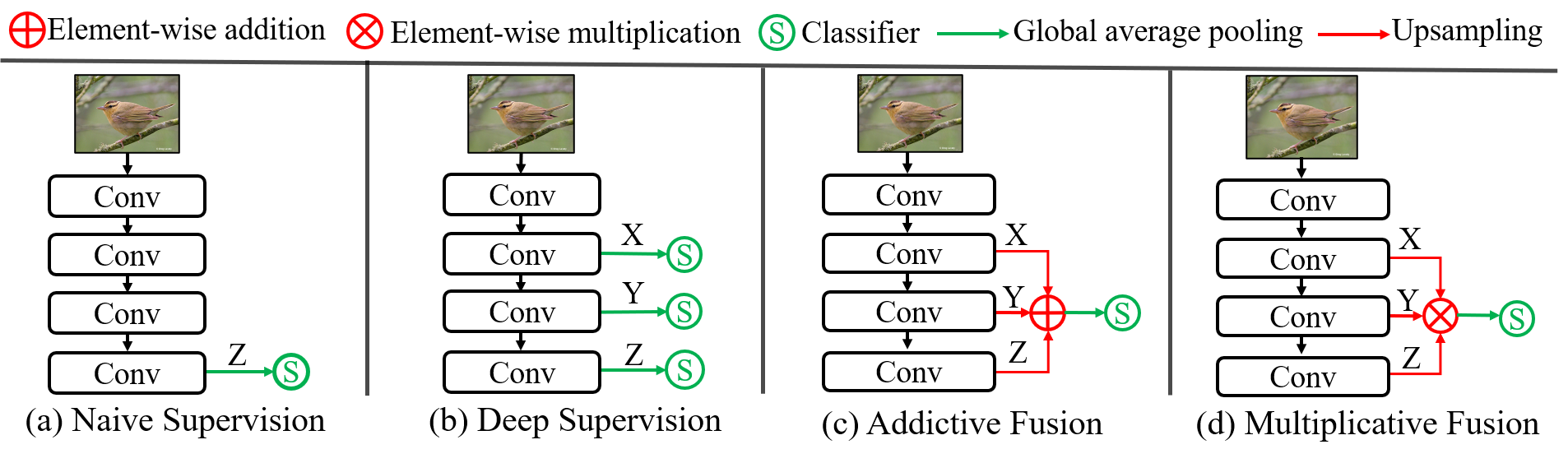}
    \caption{Different styles of supervision. In (a) and (b), deep features and combined shallow and deep features are supervised respectively. (c) and (d) show different feature fusions strategies for supervision.}
    \label{fig:supervision}
    \vspace{-0.3cm}
\end{figure*}

\section{Related Works}
\subsection{Class Activation Map (CAM) based WSOL}
Weakly supervised object localization (WSOL) is a challenging task that localizes the object only with image-level labels. For the first time, Zhou \textit{et al.}\cite{camcvpr2016} find feature maps derived from CNNs already contain object locations, even though the whole network is trained only with class labels. In view of this, they propose to replace the fully connected layer in classification models with global average pooling and utilize the class activation maps to extract object coordinates. Since the model is trained for classification, only the most discriminative parts of objects will be activated. To alleviate the problem, a lot of extensions have been proposed and remarkable progress has been made in WSOL.

\subsection{Refined CAM through Data Enhancement}
Data enhancement methods~\cite{acolcvpr2018,aecvpr2017,hideandseekiccv2017,cutmixiccv2019,eilcvpr2020,adlcvpr2019} attempt to force the model to learn from incomplete data and avoid the heavy dependency on discriminative regions. Specifically, HaS~\cite{hideandseekiccv2017} divides the input image into multiple patches. During training, only some of these patches will be used at a time so that the network will not rely on the discriminative patches too much. CutMix~\cite{cutmixiccv2019} combines the patches of two images to form a new image for training. Therefore, the network has to distinguish parts that belong to different objects. Different from random data augmentation, AE~\cite{aecvpr2017} proposes an iterative strategy to erase regions with the highest response values, repeatedly. But multiple rounds of training are computationally expensive. To improve the efficiency of AE, ACoL~\cite{acolcvpr2018} designs two branch classifiers to predict the discriminative region and corresponding complementary area at the same time. Not limited to the output, ADL~\cite{adlcvpr2019} stochastically erases multiple intermediate feature maps during forward-propagation. Through the self-attention based dropout layer, the ADL model will be enhanced for both classification and localization tasks. These erasing methods can efficiently expand object regions, but are very easy to get false positives for background regions where there is insufficient discriminative information.

\subsection{Refined CAM through Feature Enhancement}
Feature enhancement methods~\cite{spgeccv2018,daneticcv2019,psolcvpr2020,pairwiseeccv2020} try to design better mechanisms to help models learn more complete object features. Wei \textit{et al.}~\cite{mdccvpr2018} analyze the impact of object scale on predictions and propose the multi-dilated convolutional blocks (MDC) to adapt to objects with different scales. Yang \textit{et al.}~\cite{nlccamwacv2020} find all feature maps of output contribute to the final results. Rather than only using the map with the highest response, they combine all maps to suppress the background noise. Lee \textit{et al.}~\cite{ficklenetcvpr2019} consider the spatial relationships between pixels by randomly selecting hidden units. For each input image, multiple activation scores are obtained to predict the most discriminative parts. DA-Net~\cite{daneticcv2019} adopts a discrepant, divergent activation method to minimize the cosine similarity of CAMs of different branches so that each branch can learn complementary features. In~\cite{pairwiseeccv2020}, the authors argue that learning only one objective function is a weak form of knowledge transfer and propose to learn a class-wise pairwise similarity function to compare different input proposals. PSOL~\cite{psolcvpr2020} finds localization and classification interfere with each other in WSOL, which should be divided into two separate tasks, including classification and the class-agnostic localization.

All these methods have achieved great progress in WSOL. However, restrained by background noise, shallow features do not attract enough attention. In this paper, we explicitly embed shallow features into WSOL framework, and prove their vital roles for accurate object localization.

\section{Methodology}
\begin{figure*}[t]
    \centering
    \includegraphics[width=\textwidth]{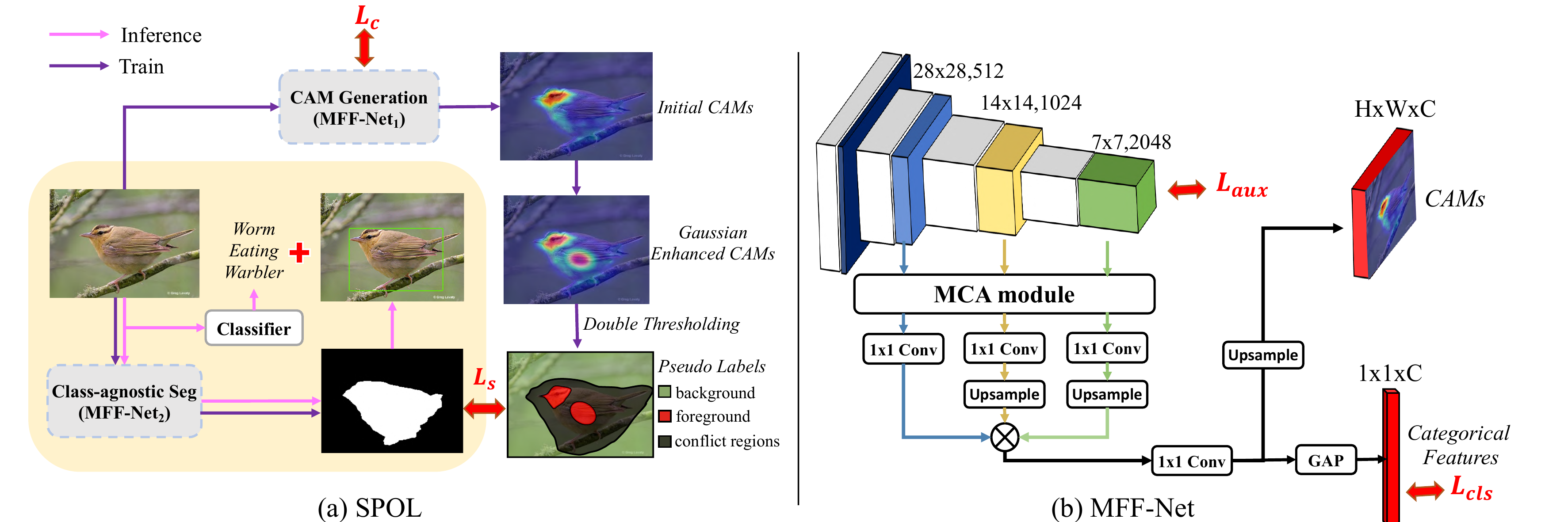}
    \caption{{\textbf{SPOL pipeline(left) and architecture of MFF-Net(right).}} (a) Pipeline of SPOL model. Purple arrows represent the training data flow. The input image firstly passes into the CAM generation module (\ie MFF-Net$_1$) to achieve the initial CAM. Then using the Gaussian distribution prior, a Gaussian enhanced CAM can be obtained, which yields the pseudo labels as the supervision for class-agnostic segmentation module (\ie MFF-Net$_2$). The inference data flow is indicated by pink arrows which is directly fed into MFF-Net$_2$ and a pre-trained classifier (\eg DenseNet161) to achieve the object mask (easily transfer to object bounding box) and the category label, respectively. 
    (b) Inner structure of MFF-Net. Feature maps from different layers (\ie shallow and deep features) are firstly processed by multiplication based channel attention (MCA) module, and then upsampled to the same resolution. Subsequently, a multiplicative feature fusion (MFF) is utilized to fuse feature at different levels, which generates the CAM and categorical features for classification.}
    \label{pipeline}
\end{figure*}

In this section, we first analyze the importance of shallow features and then elaborate our proposed SPOL model, which consists of two stages: CAM generation and class-agnostic segmentation, as shown in Fig.~\ref{pipeline}(a). 

\subsection{Rethinking Shallow Features for WSOL}
Though previous CAM-based methods have made great progress, there are still two main disadvantages.

\textbf{Coarse deep feature maps}. Due to the cascaded down-sampling operation, feature maps of the last few layers are very coarse (\eg, feature map with resolution $16\times16$ in VGG~\cite{vggiclr2014} and $8\times8$ in ResNet50~\cite{resnetcvpr2016}). 
Although such coarse feature maps do not affect the accuracy of classification, they indeed influence the object localization since the object bounding box has been degraded to a small area or even to one pixel on the coarse deep feature maps.
Enlarging the input size is an alternative way of alleviating this problem, but it brings in extra computational cost.

\textbf{Low utilization of shallow features}. For a classification task, only deep features are utilized due to their high semantics.
However, for localization task, shallow features are essential as they contain rich location information. 
Previous methods pay little attention to these shallow features since they are buried in the considerable background noise through conventional fusion strategy, as shown in Fig.~\ref{fig:teaser:framework} (a), resulting in unsatisfactory performance.

On the contrary, if the suitable strategy is utilized, shallow features can be beneficial for a better CAM generation, just as the multiplicative fusion CAMs shown in Fig.~\ref{fig:teaser:framework} (a). Concretely, shallow features can not only increase the resolution of prediction but also provide more essential details.
Thus, we propose a novel multiplicative feature fusion (MFF) to explicitly embed shallow features into deep ones.

\subsection{Multiplicative Feature Fusion}
Multi-scale feature fusion is commonly used in fully supervised high-level tasks (\ie semantic segmentation~\cite{fcncvpr2015}, object detection~\cite{fpncvpr2017}, \etc). 
However, as mentioned above, this strategy is not valid for WSOL since features from shallow layers contain too much background noise. 
Without strong supervision, detailed features are buried in these noises and do not contribute to final predictions. 
In this view, we propose the multiplicative feature fusion network (MFF-Net) to filter out the background noise of shallow features, as shown in Fig.~\ref{fig:supervision} (d). Features of different branches ({\it i.e.,} $X,Y,Z$) are firstly unsampled to the same resolution (\ie $H \times W$) then combined by element-wise multiplication for the subsequent classification head, as shown in Eq.~(\ref{eq:multiply}).
\vspace{-3pt}
\begin{align}
    \label{eq:multiply}
    F_{mul} &= \frac{1}{H \times W} \sum_{i,j}^{H,W} (X_{ij} \cdot Y_{ij} \cdot Z_{ij}) \\
    \label{eq:multiply:gradient}
    \frac{\partial F_{mul}}{\partial X_{ij}} &= \frac{1}{H \times W} Y_{ij} \cdot Z_{ij} \\
    \label{eq:addition}
    F_{add} &= \frac{1}{H \times W} \sum_{i,j}^{H,W} (X_{ij} + Y_{ij} + Z_{ij})\\
    \label{eq:addition:gradient}
    \frac{\partial F_{add}}{\partial X_{ij}} &= \frac{1}{H \times W}
\end{align}

Unlike previous methods, MFF-Net can take great advantage of shallow features since it treats the shallow and deep features in a synergistic way. 
To elaborate on this statement, we illustrate four variants with different styles of supervision in Fig.~\ref{fig:supervision}. 
Specifically, Fig.~\ref{fig:supervision} (a) is the original classification model (\eg VGG~\cite{vggiclr2014} and ResNet50~\cite{resnetcvpr2016}), where only the last layer is supervised. 
Shallow features are far from the supervision and suffer from the vanishing gradient issue. 
Fig.~\ref{fig:supervision} (b) shows the deeply supervised model~\cite{deeplyais2015}, where both deep and shallow features are directly supervised to force network to learn better representations. 
But due to the limited receptive field, shallow features have less semantics and introduce more noise. 
Thus, this direct supervision is not very helpful for WSOL. Compared with these methods, feature fusion provides a form of indirect supervision, where features of different layers are combined before the supervision. 

Fig.~\ref{fig:supervision} (c) and Eq.~(\ref{eq:addition}) show the commonly used additive fusion strategy. 
However, it does not take into consideration the correlation between multi-scale features. 
As shown in Eq.~(\ref{eq:addition:gradient}), according to chain rules, before calculating the gradients of network weights, the gradients from $F_{add}$ about each branch are the same constant without the correlation to other branches. 
That is to say, when one branch goes wrong, it does not affect other branches.
In this case, the network is not capable of learning every branch well, \eg the predictions are acceptable even the shallow features are fragmentary. 
Although it can improve the model's stability in the testing phase, it reduces the model capacity and increases the training difficulties of WSOL in the training phase.
Different from previous methods, in the proposed MFF-Net, different branches are strongly coupled through the multiplicative operation, as shown in Eq.~(\ref{eq:multiply:gradient}). 
Concretely, the gradient of $X$ branch is not constant but related to $Y$ and $Z$ branches.
These three branches will interact with each other in the training process. 
For instance, when one branch fails to capture superior representations, the multiplicative mechanism will amplify its error, and the final prediction will be wrong, leading to large gradients.
Namely, MFF-Net sets strong constraints for network training, where each branch has to learn representations well.
Furthermore, in this case, $Y$ and $Z$ are dependent on $X$ . When $X$ gets the better representations, $Y$ and $Z$ will be enhanced as a consequence. Thus, their fusion can produce more accurate predictions. 

\begin{figure}
    \centering
    \includegraphics[width=\columnwidth]{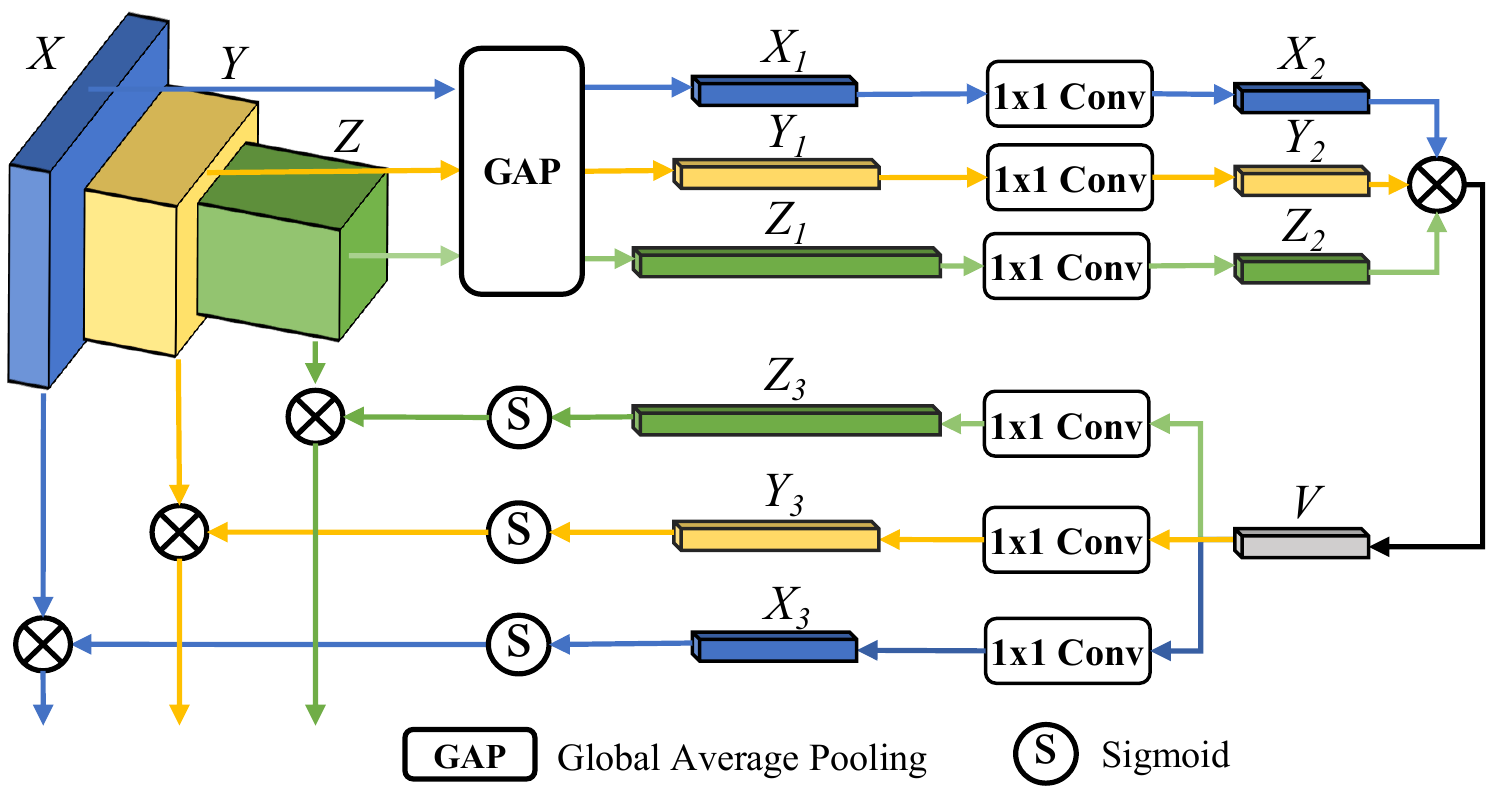}
   \caption{MCA Module. The features at different layers are firstly combined to generate a shared latent vector $V$, which will be utilized to produce the channel attention for each layer separately.}
\label{fig:channel:attention}
\end{figure}
\subsection{Multiplication based Channel Attention}
CNNs have a robust feature extraction ability, where both the foreground object and non-object background features will be represented.
For instance, some parts of the shallow features are unnecessary and can be regarded as noise, which will seriously interfere with the final predictions.
%
Thus, as shown in Fig.~\ref{fig:channel:attention}, before feature fusion of different layers, we apply channel-wise attention to roughly filter out the noise channels.
Different from traditional channel attention methods~\cite{senetcvpr2018} that only focus on one layer at a time, we propose a multiplicative channel attention (MCA) module to deal with various layers simultaneously.
Concretely, for the input feature maps $X \in \mathbb{R}^{H_1\times W_1\times C_1}$, $Y \in \mathbb{R}^{H_2\times W_2\times C_2}$ and $Z \in \mathbb{R}^{H_3\times W_3\times C_3}$, the global average pooling is first exploited to achieve $X_1 \in \mathbb{R}^{1\times C_1}, Y_1 \in \mathbb{R}^{1\times C_2}$ and $Z_1 \in \mathbb{R}^{1\times C_3}$, respectively.
Then three parallel $1\times1$ Conv layers are utilized to transfer $X_1, Y_1$ and $Z_1$ to $X_2, Y_2$ and $Z_2$ with the same shape $\mathbb{R}^{1\times C^{'}}$, respectively.
Consequently, the element-wise multiplicative fusion is used to achieve a latent representation $V=X_2 \cdot Y_2 \cdot Z_2$, and $ V\in \mathbb{R}^{1\times C^{'}}$.
Such a latent representation makes different feature representation coupled with each other. Thus, we can conduct channel attention for multiple layers at the same time.
In the reverse direction, the latent representation is transferred back to the original shape with a Sigmoid activation function, \ie $X_3 \in \mathbb{R}^{1\times C_1}, Y_3 \in \mathbb{R}^{1\times C_2}$ and $Z_3 \in \mathbb{R}^{1\times C_3}$.
Note that based on $X_3, Y_3, Z_3$, our MCA module conducts the channel attention for each corresponding layer using the multiplicative operation.

\subsection{Class-agnostic Segmentation guided WSOL}
Although MFF-Net$_1$ has produced the initial CAMs, it only focuses on the most discriminative region, which is insufficient to extract an accurate localization bounding box. 
To address this issue, we further propose the pseudo supervised class-agnostic segmentation as shown in the left sub-graph of Fig.~\ref{pipeline}, which exploits another MFF-Net$_2$. 
In this class-agnostic segmentation model, we focus on localization while discarding the category information, \ie the output only represents the foreground or background.
Since no pixel-level segmentation mask is available, we propose a two-step process to generate pseudo labels to supervise the class-agnostic segmentation module.
%
%

\textbf{Segmentation Pseudo Label Generation.} 
We complement the CAMs through a Gaussian prior pseudo label (GPPL) module for the first step.
In practice, each point $(x,y)$ on the CAMs is regarded as a sample. The response at location $(x,y)$ corresponds to its weight. 
With this setup, We calculate the mean $(\mu_x, \mu_y)$, variance $(\sigma_x^2, \sigma_y^2)$ and correlation coefficient $\rho$ between $x$ and $y$ of all samples.
Then, these parameters are applied to generate a two-dimensional Gaussian distribution, as shown in Eq.~(\ref{eq:guassian:distribution1}) and Eq.~(\ref{eq:guassian:distribution2}), which helps locate the center of object gravity and cover wide object regions, just as shown in Fig~\ref{fig:teaser:guassian} (c). 
\begin{equation}
    \label{eq:guassian:distribution1}
    \Theta=\frac{(x-\mu_x)^2}{\sigma_x^2}-2\rho\frac{(x-\mu_x)(y-\mu_y)}{\sigma_x\sigma_y}+\frac{(y-\mu_y)^2}{\sigma_y^2}
\end{equation}
\begin{equation}
    \label{eq:guassian:distribution2}
    f(x,y) = \frac{e^{-\frac{1}{2(1-\rho^2)}\Theta}}{2\pi\sigma_x\sigma_y\sqrt{1-\rho^2}}
\end{equation}
Then we ensemble the original CAMs with the Gaussian enhanced one to get a complete prediction by taking element wise maximum, just as Fig~\ref{fig:teaser:guassian} (d) shows. 
In the next step, the enhanced CAMs is further transferred into three parts with two pre-defined thresholds, as shown in the left subgraph of Fig.~\ref{pipeline}, \ie high response regions corresponding to the foreground, and low response regions corresponding to the background, and conflict regions corresponding to the areas with low confidence. 

\textbf{Class-agnostic Segmentation and Bounding Box Extraction.}
After foreground and background pseudo segmentation labels are achieved, a class-agnostic segmentation model(\ie, MFF-Net$_2$) can be trained. 
Although only a part of the image has pixel-wise labels, the segmentation model could capture a similar context and automatically cover the foreground reasonably well, illustrated in the left subgraph of Fig.~\ref{pipeline}.

After the model is well optimized, the bounding box can be extracted from the predicted mask of class-agnostic segmentation. 
The final WSOL prediction combines the extracted bounding box and the classification prediction from a standalone classifier(\eg, DenseNet161, EfficientNet-B7). 
Referring to Alg.~\ref{alg} for more details. 

\subsection{Loss Function}
For the CAM generation, apart from the classification loss, an auxiliary loss is also applied.
Specifically, two losses, \ie $\mathcal{L}_{aux}$ and $\mathcal{L}_{cls}$, are calculated through the last feature map and the fused categorical features, as shown in Fig.~\ref{pipeline} (b).
Both of these losses are calculated, using cross entropy. Thus, a joint loss $\mathcal{L}_c=\mathcal{L}_{cls}+\mathcal{L}_{aux}$ is utilized to optimize MFF-Net$_1$.
For the class-agnostic segmentation, a binary cross entropy loss is applied to supervise the segmentation model, as shown in Eq.~(\ref{eq:bce}). However, only the pseudo foreground and background regions are considered except for the areas of conflict. Concretely, for the foreground and background, $w_{ij}$ equals to 1, while for those conflict regions, $w_{ij}$ is set to zero. In this way, the losses of conflict area are ignored to avoid misleading the network.
\begin{small}
\begin{equation}
    \label{eq:bce}
    \mathcal{L}_{seg}=-\frac{\sum\limits_{(i,j)}w_{ij}[g_{ij}log(p_{ij})+(1-g_{ij})log(1-p_{ij})]}{H \times W}
\end{equation}
\end{small}
where $p_{ij}$ and $g_{ij}$ are the predicted probability and ground truth label at position (i,j), respectively.

%

\SetKwInOut{Initialization}{Initialization}
\begin{algorithm}[!t]
        \label{alg}
        \caption{\small SPOL}
        \LinesNumbered 
        \KwIn{Training images $I_{tr}$ with class label $L_{tr}$}
        \KwOut{Predicted bounding boxes $B_{te}$ and class labels $L_{te}$ on testing images $I_{te}$}
        \BlankLine 
        // Training Phase\\
        Train MFF-Net$_1$ $F_w$ on $I_{tr}$ with $L_{tr}$\\
        Use $F_{w}$ to generate pseudo label ${M}_{tr}$ on $I_{tr}$ \\
        Train MFF-Net$_2$ $F_{s}$ on $I_{tr}$ for Seg. with ${M}_{tr}$\\
        Train a classifier $F_{c}$ on $I_{tr}$ with $L_{tr}$\\
        
        \BlankLine
        // Inference Phase\\
        Use $F_{s}$ to predict $M_{te}$ on $I_{te}$\\
        Extract object bounding box $B_{te}$ from $M_{te}$\\
        Use $F_{c}$ to predict $L_{te}$ on $I_{te}$\\
        \textbf{Return:} $B_{te},L_{te}$
\end{algorithm}

\section{Experiments}

\subsection{Experimental Setup}
\textbf{Datasets.} To evaluate the proposed SPOL, two datasets are adopted, including CUB-200~\cite{cubtech2011} and ImageNet-1K~\cite{imagenetijcv2015}. CUB-200 contains 200 categories of birds with 5,994 training images and 5,794 testing images. ImageNet-1K is a much larger dataset with 1000 classes, containing 1,281,197 training images and 50,000 validation images.
%

\textbf{Metrics.} Following previous methods~\cite{camcvpr2016,adlcvpr2019}, three metrics are adopted for evaluation. 1) Top-1 localization accuracy (\textit{Top-1 Loc}): fraction of images with right prediction of class label and more than 50\% IoU with the ground-truth box. 2) Top-5 localization accuracy (\textit{Top-5 Loc}): fraction of images with class labels belonging to Top-5 predictions and more than 50\% IoU with the ground-truth box. 3) GT-known localization accuracy (\textit{GT-known Loc}): fraction of images for which the predicted bounding box has more than 50\% IoU with the ground-truth box.

%

\subsection{Implementation Details}
\textbf{Network Architecture.} We adopt the pre-trained ResNet50~\cite{resnetcvpr2016} as our backbone network for MFF-Net. 
Considering the MFF-Net can generate the CAMs and categorical features simultaneously, two separate MFF-Nets are trained for the CAM generation and class-agnostic segmentation.
Besides, during inference, for the extra classifier illustrated in the Fig.~\ref{pipeline} (a), DenseNet161~\cite{deeplyais2015} is exploited to predict image class for a fair comparison.


\textbf{Training Setting and Data Preprocessing.} On CUB-200 dataset, we train 32 epochs for two MFF-Nets. 
%
The learning rate always keeps the same, about 0.002 for the pre-trained feature extractor and 0.02 for newly added layers.
On ImageNet-1K, we also utilize pre-trained weights and the same learning rates, but the number of training epochs is set as 6 correspondingly. 
In the training phase, we first resize the input image to $256\times256$ then randomly crop it to $224\times 224$. Besides, a random flip is also adopted to augment input images. In the testing phase, we replace the random cropping with the center cropping as previous works~\cite{psolcvpr2020,adlcvpr2019}. During Gaussian enhancement, we regard values large than 0.7 as the foreground. For the pseudo label generation after Gaussian enhancement, the double threshold for the foreground and background is set as 0.5 and 0.004, respectively.

\subsection{Comparison with state-of-the-arts}
\begin{table*}[htb]
	\caption{Comparison with state-of-the-art methods. 'Cls Backbone' is the network used for classification and 'Loc Backbone' represents the network used for localization. '-' means that the authors do not provide corresponding results. Best results are highlighted in bold.}
	\label{tab:compall}
	\setlength{\tabcolsep}{4pt}
	\small
	\centering
	\begin{tabular}{l|c|c|c|c|c|c|c|c}
		\hline
		\hline
		\multirow{2}{*}{Model} & \multirow{2}{*}{Loc Backbone} & \multirow{2}{*}{Cls Backbone} & \multicolumn{3}{c}{CUB-200}&\multicolumn{3}{|c}{ImageNet-1K}\\ 
		\cline{4-6} \cline{7-9}
		& &  & Top-1 Loc & Top-5 Loc & GT-Known Loc & Top-1 Loc &Top-5 Loc & GT-Known Loc\\
		\hline 
		\hline 
		CAM~\cite{camcvpr2016} & \multicolumn{2}{c|}{VGG-GAP} & 36.13 & - & -  & 42.80 & 54.86 & 59.00\\
		ACoL~\cite{acolcvpr2018} &
		\multicolumn{2}{c|}{VGG-GAP}  & 45.92 & 56.51& 62.96  & 45.83 & 59.43 & 62.96\\
		ADL~\cite{adlcvpr2019} & \multicolumn{2}{c|}{VGG-GAP}  & 52.36 & -& 73.96 & 44.92 & - & -\\
		DDT~\cite{ddtpr2019} &  \multicolumn{2}{c|}{VGG16}  & 62.30 & 78.15&  84.55 & 47.31 & 58.23 & 61.41\\
		SPG~\cite{spgeccv2018} &  \multicolumn{2}{c|}{InceptionV3} & 46.64 & 57.72& - & 48.60 & 60.00 & 64.69\\
		ADL~\cite{adlcvpr2019} &  \multicolumn{2}{c|}{ResNet50-SE} & 62.29 & - & 71.99 & 48.53 & - & -\\
		ADL-TAP~\cite{bae2020rethinking} &  \multicolumn{2}{c|}{GoogleNet} & 53.04 & - & 69.95 & 50.56 & - & 64.44\\
		$I^2C$~\cite{zhang2020inter} &  \multicolumn{2}{c|}{InceptionV3} & 65.99 & 68.34 & 72.60 & 53.11 & 64.13 & 68.50\\
		GC-Net~\cite{lu2020geometry} &  \multicolumn{2}{c|}{GoogLeNet} & 58.58 & 71.10 & 75.30 & 49.06 & 58.09 & -\\
		\hline
		\hline
		PSOL\cite{psolcvpr2020} & InceptionV3 & InceptionV3  & 65.51 & 83.44& - & 54.82 & 63.25 & 65.21 \\
		PSOL\cite{psolcvpr2020} &  ResNet50& ResNet50  & 70.68 & 86.64& 90.00 & 53.98 & 63.08 & 65.44
		\\
		PSOL\cite{psolcvpr2020} & DenseNet161 &  DenseNet161 & 74.97 & 89.12 & 93.01  & 55.31 & 64.18 & 66.28 \\
		PSOL\cite{psolcvpr2020} & DenseNet161 & EfficientNet-B7 & 77.44 & 89.51 & 93.01 & 58.00 & 65.02 & 66.28
		\\
		\hline
		SPOL(Our) & ResNet50 & DenseNet161 & 79.74 & \textbf{93.69} & \textbf{96.46} & 56.40 & 66.48 & \textbf{69.02}\\
		SPOL(Our) & ResNet50 & EfficientNet-B7 & \textbf{80.12} & 93.44 & \textbf{96.46} & \textbf{59.14} & \textbf{67.15} & \textbf{69.02}\\
		\hline
		\hline
	\end{tabular}
\end{table*} 
\textbf{Quantitative Comparison} To demonstrate the effectiveness of the proposed SPOL, we compare against previous methods ~\cite{adlcvpr2019, cutmixiccv2019,ddtpr2019,spgeccv2018,camcvpr2016,psolcvpr2020} in terms of \textit{Top-1 Loc}, \textit{Top-5 Loc} and \textit{GT-Known Loc}, which is shown in Tab.~\ref{tab:compall}.
%
Best results are highlighted in bold. 
The proposed SPOL outperforms previous state-of-the-art methods by a large margin, especially on the practical criterion \textit{GT-Known Loc} metric.
%
Among all the variants, SPOL achieves the highest accuracy on both CUB-200 and ImageNet-1K.

\begin{figure*}[htb]
    \centering
    \includegraphics[width=\textwidth]{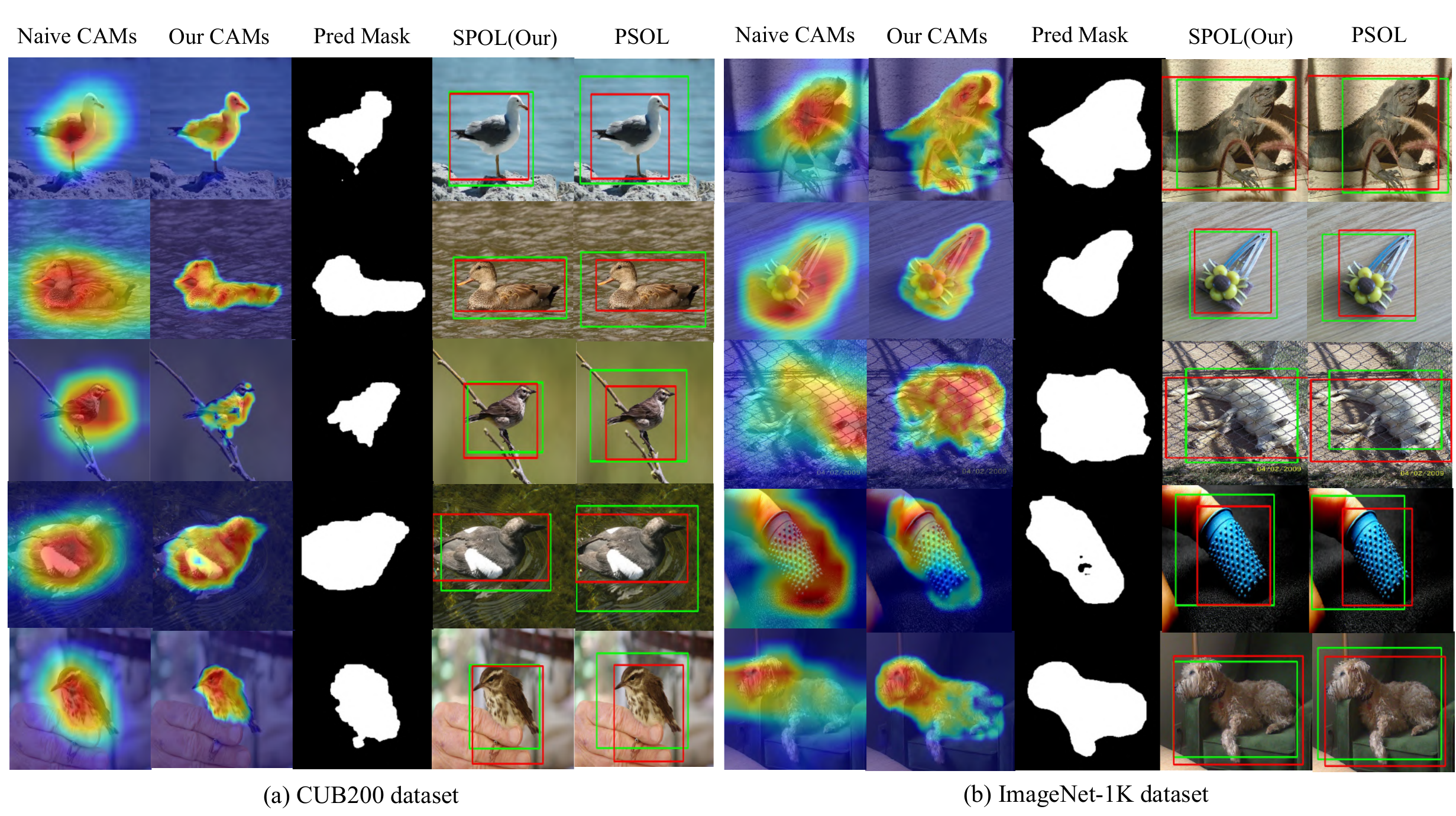}
    \caption{Visual comparisons of different methods. The 1st column shows the naive CAMs. The 2nd column shows the CAMs produced by MFF-Net. The 3rd column shows the predicted mask by our proposed class-agnostic segmentation model. The last two columns show the ground truth bounding box (red color) and the predicted bounding box (green color) of our model and previous PSOL~\cite{wei2019unsupervised}, respectively.}\label{fig:viscmp}
    \vspace{-0.5cm}
\end{figure*}
\textbf{Visualization Comparison} Visualization comparisons of the proposed SPOL and other methods ({\it CAM~\cite{camcvpr2016} and PSOL~\cite{wei2019unsupervised}}) are shown in Fig.~\ref{fig:viscmp}. 
From the first two columns, it is evident that CAMs produced by our proposed MFF-Net has much sharper boundaries than naive CAMs, regardless of CUB200 or ImageNet-1K dataset. 
Besides, our CAMs can cover more complete object regions rather than only focus on the most discriminative ones. 
It exactly proves that the proposed MFF-Net benefits from multiplicative feature fusion and Gaussian enhancement. 
The third column shows the class-agnostic segmentation model's predicted mask, which further refines the object regions. 
The last two columns show the predicted bounding box (green color) and the ground-truth one (red color).
Bounding boxes produced by our SPOL not only localize object regions accurately but also are more compact than previous PSOL~\cite{psolcvpr2020}, which verifies the superiority of SPOL.

\subsection{Ablation Study} 
In this section, we replace or remove the specific components in our proposed model to match with 'their importance' and conduct evaluation on CUB200 dataset. 

\textbf{Results with different fusion strategies.} 
\begin{table}[t]
  \caption{Model performance with the number of involved shallow features, evaluated by \textit{GT-known Loc}. Fuse$X$ means that features of last $X$ stages of ResNet50 are aggregated. }
  \label{fuseX}
  \setlength{\tabcolsep}{8pt}
  \centering
  \begin{tabular}{l|cccc}
    \hline
    \hline
    Method          & Fuse1 & Fuse2 & Fuse3 & Fuse4  \\
    \hline
    Addition        & 55.76 & 55.19 & 55.00 & 55.10 \\
    Concatenation   & 55.76 & 53.14 & 53.40 & 53.32 \\
    Multiplication  & 55.76 & {\textbf{79.05}} & {\textbf{88.30}} & {\textbf{84.04}} \\
    \hline
    \hline
  \end{tabular}
\end{table}
We compare the model performance under different fusion strategies ({\it i.e.,} addition, concatenation and multiplication). 
As shown in Tab.~\ref{fuseX}, addition or concatenation based fusion does not improve with more shallow features involved (from Fuse1 to Fuse4). 
On the contrary, multiplicative fusion methods achieve significant improvements, demonstrating their superiority and the importance of shallow features for WSOL. 
Besides, as shown in the Tab.~\ref{fuseX}, fusing the features from the last three layers achieves the highest performance, which is exactly the applied setting in our model.

\begin{figure}[t]
    \centering
    \includegraphics[width=\columnwidth]{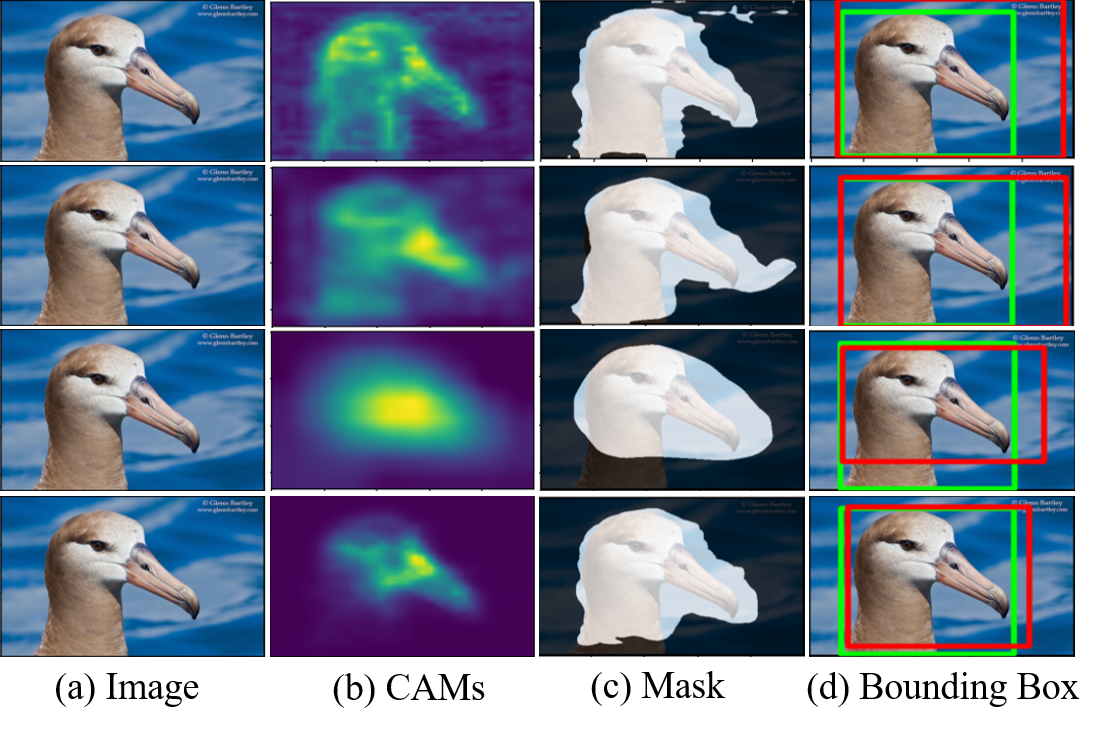}
   \caption{Predictions of different layers. (a) shows the input image. (b) shows the predicted CAMs. (c) derives the mask from (b) by binarization. (d) shows the predicted bounding box (red color) and ground truth bounding box (green color). The above three rows represent predictions of different layers, and the last row shows the predictions using multi-layer fusion.}
    \label{fig:branch}
\end{figure}

\textbf{Visualization of different layers from MFF-Net.} 
%
As shown in Fig.~\ref{fig:branch}, the first three rows represent the predictions of the shallow, middle, and deep layers of MFF-Net, respectively, and the last row shows the predictions of the fused features. We can see shallow features (first row) have much sharper boundaries than deep ones (third row), while deep features contain less background noise than shallow ones. Combining all features (fourth row) could predict a more accurate object mask and bounding box.

\textbf{Ablation studies of MFF-Net components.} We further compare the effectiveness of each component proposed in MFF-Net, as shown in Tab.~\ref{abla2}. Precisely, we measure the performance reduction by removing multiplicative channel-wise attention (MCA) and auxiliary loss (Aux), respectively. Once MCA module is removed, \textbf{GT-Known Loc} falls from 92.25\% to 88.52\%, which proves the importance of MCA in noise channel removal. The auxiliary loss is also necessary for MFF-Net regarding the performance drops, which helps the model improve convergence.

\begin{table}[t]
    \label{abla2}
	\caption{Ablation studies of MFF-Net's components. `w/o MCA Module` indicates that features are directly fused without channel-wise attention. `w/o Aux Loss` means that auxiliary loss is abandoned.}
	\setlength{\tabcolsep}{2pt}
	\centering
	\begin{tabular}{l|c|c|c}
		\hline
		\hline
		Model  & Top-1 Loc & Top-5 Loc & GT-Known Loc\\  
		\hline 
		w/o Aux Loss & 73.42 & 86.21 & 89.45\\
		w/o MCA Module & 72.83 & 85.55 & 88.52\\
		MFF-Net(Our) & {\textbf{75.82}} & {\textbf{89.13}} & {\textbf{92.25}}\\
		\hline
		\hline
	\end{tabular}
  \vspace{-0.3cm}
\end{table}

\textbf{Ablation studies of class-agnostic segmentation model.}
We conducted ablation studys to further illustrate the improvement of different components for the class-agnostic segmentation model. 
%
As shown in Tab.~\ref{tab:abla1}, \textbf{GT-Known Loc} accuracy in `w/o Seg` line decreases from 96.46\% to 92.25\% when the class-agnostic segmentation model is not applied.
It shows that the segmentation model improves the consistency of predictions and produces more completed object regions. 
Besides, the quality of pseudo labels is crucial for the segmentation model.
%
As shown in Tab.~\ref{tab:abla1}, either removing the thresholding or Gaussian enhancement step hinders the segmentation performance. 

\begin{table}
	\caption{Ablation studies for class-agnostic segmentation model on CUB-200 dataset. `w/o Seg` indicates removal of the segmentation model and directly outputting bounding boxes from CAM generation stage. `w/o Threshold` indicates no thresholding is performed on initial CAMs. The pseudo label is generated subjectively by Gaussian enhancement. `w/o Gauss Enhance` indicates only thresholding step is applied without Gaussian enhancement.}
	\label{tab:abla1}
	
	\setlength{\tabcolsep}{2pt}
	\centering
	\begin{tabular}{l|c|c|c}
		\hline
		\hline
		Model  & Top-1 Loc & Top-5 Loc & GT-Known Loc\\  
		\hline 
		w/o Threshold     & 60.39 & 71.38 & 73.97\\
		w/o Gauss Enhance & 78.08 & 91.82 & 95.41\\
		w/o Seg           & 75.82 & 89.13 & 92.25\\
		SPOL(Our)  & {\textbf{78.94}} & {\textbf{92.84}} & {\textbf{96.46}}\\
		\hline
		\hline
	\end{tabular}
	\vspace{-0.5cm}
\end{table}

\section{Conclusion}
In this paper, we propose the {\textbf{S}}hallow feature-aware {\textbf{P}}seudo supervised {\textbf{O}}bject {\textbf{L}}ocalization (SPOL) model for accurate WSOL. 
We first analyze the importance of shallow features for object detection and then show that conventional fusion ignores the power of shallow features due to background noise interference. 
Thus, a multiplicative feature fusion strategy is introduced to utilize shallow features, suppress background noise and enhance object boundaries.
Further, a class-agnostic segmentation model is trained with the pseudo labels to refine object predictions.
Extensive experiments verify the effectiveness of the proposed SPOL, which outperforms previous methods by a large margin.

\section{Acknowledgement}
The work was supported in part by NSFC-Youth 61902335, by the Key Area RD Program of Guangdong Province with grant No. 2018B030338001, by the National Key RD Program of China with grant No. 2018YFB1800800, by Guangdong Regional Joint Fund-Key Projects 2019B1515120039, by Shenzhen Outstanding Talents Training Fund, by Guangdong Research Project No. 2017ZT07X152 and by CCF-Tencent Open Fund.

\newpage
{\small
\bibliographystyle{ieee_fullname}
\bibliography{egbib}
}
\end{document}